\documentclass{article}
\usepackage{kr}

\usepackage{times}
\usepackage{soul}
\usepackage{url}
\usepackage[hidelinks]{hyperref}
\usepackage[utf8]{inputenc}
\usepackage[small]{caption}
\usepackage{graphicx}
\usepackage{amsmath}
\usepackage{amsthm}
\usepackage{booktabs}
\usepackage{algorithm}
\usepackage{algorithmic}
\usepackage{paralist}
\usepackage{amsmath}
\usepackage{amssymb}
\usepackage{paralist}
\usepackage{tabularx}
\usepackage{xspace}
\usepackage{multirow}   
\usepackage{subcaption}
\usepackage{makecell}
\usepackage{multicol}
\usepackage{float}
\usepackage{cuted}    
\usepackage{listings}
\usepackage{microtype}
\usepackage{color}
\usepackage[dvipsnames]{xcolor}

\newcommand{\iec}{i.e.,\xspace}
\newcommand{\egc}{e.g.,\xspace}

\newcommand{\theory}{theory\xspace}

\newcommand{\pipelinetitle}{GS-VQA\xspace}
\newcommand{\asplarrow}{\mathrel{\mathrm{:\!\!{-}}}}

\AddToHook{cmd/ttfamily/after}{\frenchspacing}

\newcommand{\citeNBYB}[1]{\citeauthor{#1}~\shortcite{#1}}

{\end{list}}

\newcounter{myenumctr}

\newcommand{\nop}[1]{}

\newcommand{\myparagraph}[1]{

\smallskip\vspace{2pt} 

\noindent{\bf#1}~}

\title{Declarative Knowledge Distillation from \\Large Language Models for Visual Question Answering Datasets}

\author{%
Thomas Eiter$^1$\and
Jan Hadl$^1$\and
Nelson Higuera$^1$\and
Johannes Oetsch$^2$\\
\affiliations
$^1$Institute for Logic and Computation, TU Wien, Favoritenstraße 9--11, 1040 Vienna, Austria\\
$^2$Department of Computing, Jönköping University, Gjuterigatan 5, 551$\,$11 Jönköping, Sweden\\
\emails
\{thomas.eiter, 
jan.hadl, 
nelson.ruiz\}@tuwien.ac.at,
johannes.oetsch@ju.se
}
\begin{document}

\maketitle

\begin{abstract}
Visual Question Answering (VQA) is the task of answering a question about an image and requires processing multimodal input and reasoning to obtain the answer. Modular solutions that use declarative representations within the reasoning component have a clear advantage over end-to-end trained systems regarding interpretability. The downside is that crafting the rules for such a component can be an additional burden on the developer. We address this challenge by presenting an approach for declarative knowledge distillation from Large Language Models (LLMs). Our method is to prompt an LLM to extend an initial theory on VQA reasoning, given as an answer-set program, to meet the requirements of the VQA task. Examples from the VQA dataset are used to guide the LLM, validate the results, and mend rules if they are not correct by using feedback from the ASP solver. We demonstrate that our approach works on the prominent CLEVR and GQA datasets. Our results confirm that distilling knowledge from LLMs is in fact a promising direction besides data-driven rule learning approaches.%
\end{abstract}

\section{Introduction}\label{sec:intro}

\emph{Visual question answering} (VQA)~\cite{antol2015vqa,goyal2017vqav2} is a challenging problem with valuable applications~\cite{barra2021vqaapps,DBLP:journals/artmed/LinZTSHWHG23}; it is the  task of providing an accurate answer for a question about a visual scene. This requires not just a joint understanding of vision and text, but also the ability to follow complex chains of reasoning operations.

\emph{Neurosymbolic approaches to VQA}~\cite[etc.]{MaoGKTW19,yi2019neuralsymbolicvqa,amizadeh2020neurosymbolic,eiter2022neuro,vipergpt,JohnstonNS23a} use deep learning for perception, produce a symbolic representation of the input image and question, and then perform reasoning on this representation in a purely symbolic way. 
These approaches are interpretable, transparent, and can be extended easily due to their compositional structure. 
%
A promising direction in this regard is to use logic-based formalisms for the reasoning component. 
We are in particular interested in using Answer-Set Programming (ASP)~\cite{brewka2011asp,lifschitz2019answer}, a prominent knowledge representation framework, for the reasoning module of such systems. Besides concise representations, advantages are that 
non-determinism allows for multiple answers if desired,
ambiguity in the perception often can be resolved in the reasoning module~\cite{eiter2022neuro}, and one can more easily add explanation capabilities~\cite{EiterGHO23}.  
Using ASP to augment VQA with reasoning capabilities is a topic that is indeed currently gaining traction and is used not only for VQA~\cite{AbrahamAR24,EiterGHO23,eiter2022neuro,BasuSG20}, but also for tasks such as
segmentation of medical images~\cite{brunoCMM21}.
%
The downside is that crafting the rules for such a component is not always easy and can be an additional burden on the developer.

\begin{figure}[!t]
    \centering
    \includegraphics[width=\linewidth]{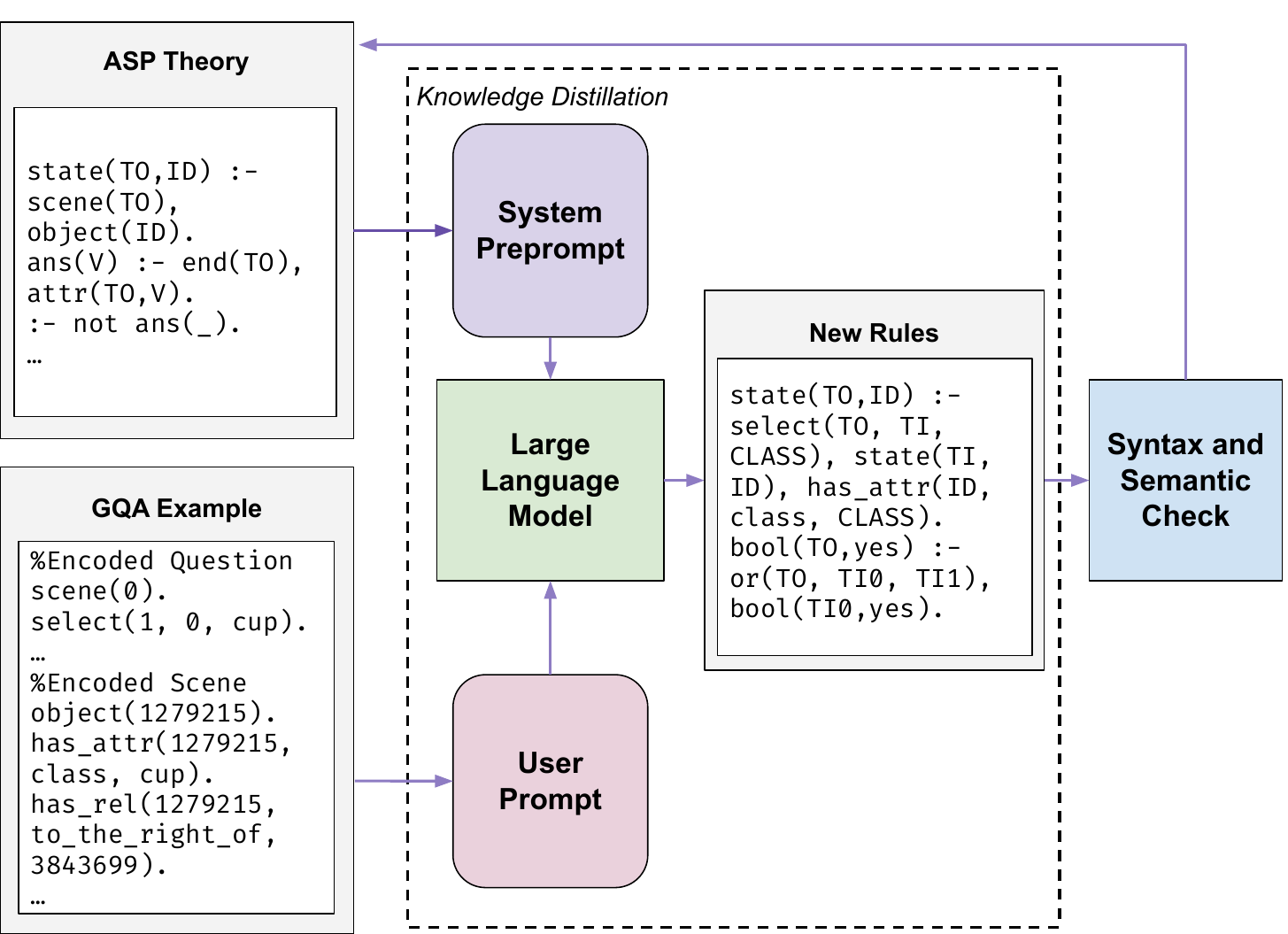}
    \caption{An overview of our knowledge distillation method.}
    \label{fig:knowledge_distillation}
\end{figure}

We address this challenge by presenting an approach for \emph{declarative knowledge distillation} from Large Language Models (LLMs)~\cite{NIPS2017_3f5ee243,DBLP:journals/corr/abs-2303-18223} (see Fig.~\ref{fig:knowledge_distillation}). The premise of this work is that we want to develop a modular solution for a VQA task for which we have, as is commonly the case, a dataset of questions, scenes, and answers at our disposal. Further, the reasoning component involves already an initial ASP theory that covers some but not all aspects of the VQA task. Our method is to prompt an LLM to extend this initial theory to meet the remaining requirements. Examples from a VQA dataset are used to guide the LLM, validate the results, and mend the generated rules if they are not correct using feedback from the ASP solver.
More specifically, if we encounter an example from the dataset for which the system does not yet give the correct answer, we prompt the LLM with a description of the scene, question, and expected answer to suggest rules to extend the current program. We then run an ASP solver to test the new program. If the output of the ASP solver does not match the expected answer, we hand this ouput and the expected answer to the LLM and prompt it to correct the ASP program accordingly. If the ASP solver gives a syntax error, we ask the LLM to fix it.
If the test passes, we proceed with regression testing on previous examples to ensure that the system will still answer any question it could answer before the extension. 
We also experiment with presenting examples in batches to reduce the number of calls to the LLM.

Our approach bears similarities with methods from statistical-relational learning (SRL)~\cite{RaedtK17}, as there the goal is also to find rules that generalise examples from a dataset. While SRL is data-driven, our method uses examples only to guide a knowledge distillation process via conditioning an LLM; we go into more detail in the related work section.

We use two VQA datasets to evaluate our knowledge distillation method: CLEVR~\cite{johnson2017clevr} and GQA~\cite{hudson2019gqa}. CLEVR uses synthetic scenes and is designed to challenge VQA systems with compositional questions that require breaking down the overall task into a tree of primitive operations that is evaluated recursively. On the other hand, GQA uses real images that depict complex visual scenes and diverse questions with a large number of possible answers that involve reasoning on the image scene graph, \iec the objects in the scene, their attributes, and relations. To solve CLEVR, we use a recent VQA system~\cite{EiterGHO23}. For GQA, we use a similar system~\cite{Hadl23}, which is able to perform VQA in a \emph{zero-shot manner}, \iec without training or fine-tuning of components to the current dataset.
This is a significant advantage over related systems that require a costly training process. Both VQA systems use ASP for reasoning.

Our experiments show that LLMs can understand and produce ASP rules to implement new reasoning operators, and thus can effectively assist in the process of generating the reasoning component. 
This methodology offers a promising avenue for extracting domain-specific knowledge from LLMs in the realm of VQA and offers a viable alternative to data-driven rule learning approaches.

The remainder of the paper is organised as follows. We discuss related work in Section~\ref{sec:rel} and present the VQA approaches used in Section~\ref{sec:gsvqa}. Then, we introduce the knowledge distillation method in Section~\ref{sec:distillation} and evaluate it in Section~\ref{sec:experiments}. Finally, Section~\ref{sec:concl} presents
concluding remarks and pointers to future work.

\section{Related Work}\label{sec:rel}

\noindent
\emph{Answer-Set Programming and LLMs.$\quad$}
Recently, LLMs have been explored in the context of ASP. \citeNBYB{ishay2023llmasp} observed that LLM reasoning capabilities are shallow, but they can serve as a highly effective semantic parser to transform input into ASP representations. These are then used to solve logical puzzles.
We also recently proposed to use LLMs to parse the question into ASP facts in the context of VQA for images of graphs~\cite{bauer2023}.
\citeNBYB{gupta2023reliable} showed that the combination of LLMs with ASP in their STAR framework produces good results for natural language tasks that require qualitative reasoning, mathematical reasoning, and goal-directed conversation. 
Our work is different as we do not focus on semantic parsing but on the more challenging task of knowledge distillation, where we aim for a system that can produce sound logical rules capturing knowledge about a particular domain.
Recent work by \citeNBYB{zhu2023llm} explores the use of LLMs to learn rules from arithmetic and kinship relationships, yet the rules they learn do not contain variables and their semantics is informal.

\smallskip
\noindent
\emph{Statistical-relational learning.$\quad$}
Similar to our approach, methods from statistical-relational learning (SRL)~\cite{RaedtK17}, in particular from inductive logic programming (ILP)~\cite{DBLP:journals/ngc/Muggleton91,MuggletonR94,DBLP:journals/ml/CropperDEM22}, aim at producing rules from example data and a background theory.
SRL has seen great advances in terms of scalability by, \egc applying gradient-based boosting~\cite{GutmannK06}, and systems like ILASP~\cite{law20} and FastLAS~\cite{LawRBB020} provide means for inductively learning expressive ASP programs. 

However, SRL takes a statistical and probabilistic learning perspective that is in essence data driven. Our method is orthogonal to that,  as it does not aim at learning. On the surface, we also use a data set, but the role is very different as it guides a knowledge distillation process via conditioning an LLM. 
ILP uses a search-based approach where the solutions produced are correct and minimal under some criteria. A key aspect of many ILP systems are {\em mode declarations} that define the syntactic form of allowed rules to restrict the search space of possible programs. Mode declarations are in a 
formal language, and they tacitly assume an intuition about the form of the solution.
For the distillation approach , we do not need that; we only elicit knowledge that is already present in the LLM, and the information in the prompt that instructs what rules we want is informal and in natural language. When prompting LLMs, rules are general by command---while optimality is not enforced, it may happen implicitly.

\smallskip
\noindent
\emph{Modular neurosymbolic VQA.$\quad$}
There are several VQA systems that feature a modular architecture which combines subsymbolic with symbolic components~\cite{yi2019neuralsymbolicvqa,MaoGKTW19,amizadeh2020neurosymbolic,eiter2022neuro,vipergpt,JohnstonNS23a}.
Specifically, \citeNBYB{yi2019neuralsymbolicvqa} used a pipeline to extract a \emph{scene graph} (a list of all objects detected in the image with their attributes and positions) from the image. They then translated the provided question into a structured representation of the reasoning steps, called \emph{functional program}, and executed this program on the structural scene representation to obtain an answer. The authors showed excellent results on the popular CLEVR dataset  ~\cite{johnson2017clevr}. 
This approach has been advanced with logic-based reasoning processes, \egc by Differentiable First-Order Logic ($\nabla$-FOL)~\cite{amizadeh2020neurosymbolic}, or by ASP~\cite{eiter2022neuro}. 
These reasoning processes 
can consider not just the most probable scene-graph prediction, but rather the entire vector of probabilities as output by the object detection and attribute/relation classifier networks that form the visual perception component of the VQA pipeline.
%
Foundational models such as Vision-Language Models (VLMs) such as BLIP-2~\cite{li2023blip2} and SimVLM~\cite{wang2022simvlm}  have become sufficiently strong through their pre-training regimes to generalise well to multiple different datasets. Approaches that use these VLMs as components are, \egc ViperGPT~\cite{vipergpt}, CodeVQA~\cite{subramanian2023codevqa}, and PnP-VQA~\cite{tiong2023pnpvqa}.

\section{Background \& VQA Methodology}\label{sec:gsvqa}

In this section, we review the basics of the logic-based VQA approaches for the two datasets, GQA and CLEVR, that we are going to use for our evaluation. Both systems use ASP to derive answers from a symbolic scene representation. So we start by reviewing the basics of ASP next.

\subsection{Answer-Set Programming}
    
Answer-Set Programming (ASP)~\cite{brewka2011asp,lifschitz2019answer}
is a well-known approach to declarative problem solving, in which 
solutions to a problem are described by sets of logical rules. Efficient ASP solvers for evaluating the rules are readily available.\footnote{We use clingo (v.~5.6.2 ) from
\url{https://potassco.org/.}} 

For our concerns, an ASP program is a finite set $P$ of rules $r$ of the 
following form:
{\begin{align}\label{eq:rule}
a \asplarrow\ b_1,
\ldots,\ b_n,\ not\ c_1, \ldots,\ \mathit{not}\ c_n\nonumber \quad m,n \geq 0\,
\end{align}}%
where $a$, all $b_i$, and all $c_j$ are atoms in a
first-order predicate language, and $\mathit{not}$ stands for negation
as failure (aka.\ weak negation). We allow that $a$ may be missing (viewed as falsity);
then $r$ acts as a constraint. 
Intuitively, the rule means that whenever all $b_i$ are true and none
of the $c_j$ can be shown to be true, then $a$ must be true.  
Some rules appear in Fig~\ref{fig:encodings}. 

The semantics of a ground (variable-free) ASP program is given in terms of
answer sets, which are Herbrand models that satisfy a stability
condition~\cite{GelfondL88}.  A Herbrand interpretation of $P$ is a set $I$ of ground
atoms in the language induced by $P$ (intuitively, the atoms that are
true). Such an $I$ is a model of $P$ if for each rule $r$ in $P$ either
$(i)$ $a \in I$ or $(ii)$  $\{b_1,\ldots,b_n\}\not\subseteq I$ or
$(iii)$  $I\cap \{c_1,\ldots,c_n\}\neq \emptyset$; that is, $I$ satisfies $r$
viewed as implication in classical logic.

An interpretation $I$ is then an answer set of $P$,
if  $I$ is a $\subseteq$-minimal model of the program $P^I = \{ r \in
P \mid I$ satisfies neither $(ii)$ nor $(iii) \}$. Intuitively, 
$I$ must result by applying the rules $r$ whose bodies
``fire'' w.r.t.\ $I$ starting from facts.

The semantics of programs with variables is defined in terms of their
groundings (uniform replacement of variables in rules with all
possible ground terms).

ASP features further constructs such as choice rules (which allows to
select among alternatives under cardinality bounds) and weak
constraints (\iec soft constraints expressing costs for an
objective function that is 
minimised); notably, the latter
allow for modeling numeric uncertainty and to single out the most
likely from answer sets of a program or a range of most likely answer
sets. For more details on ASP, we refer to 
\cite{brewka2011asp,calimeri2012asp}.

\subsection{Zero-Shot VQA for the GQA Dataset}

Next, we explain our system for zero-shot VQA for the GQA dataset~\cite{hudson2019gqa}. It does not require any training, but relies on foundation models for processing the visual scene and ASP for deducing answers. 

\subsubsection{The GQA dataset.}

We use the state-of-the-art GQA dataset, which has been widely adopted in the recent literature \cite{amizadeh2020neurosymbolic,vipergpt,liang2020lrta,li2023blip2}. 
It contains over 22M open and binary questions that are complex in structure, involve a wide variety of reasoning skills, and have a large number ($1\,878$) of possible answers.  
The questions cover more than $100\,000$ images from the Visual Genome dataset \cite{krishna2016visualgenome} that present real-world scenes with a wide variety of object classes, attributes, and relations. 
GQA comes with two types of supplementary data that greatly aid in the development of our pipeline: 
First, each natural-language question from the test split comes with a functional representation of its required reasoning steps.
Second, a Visual Genome scene graph is provided for every image in the test split of the dataset, which allows us to verify soundness of our ASP encoding under perfect visual information.

\subsubsection{Solving GQA.}
\begin{figure*}[!t]
\centering
\begin{subfigure}{.7\textwidth}
  \centering
    \includegraphics[width=0.9\textwidth]{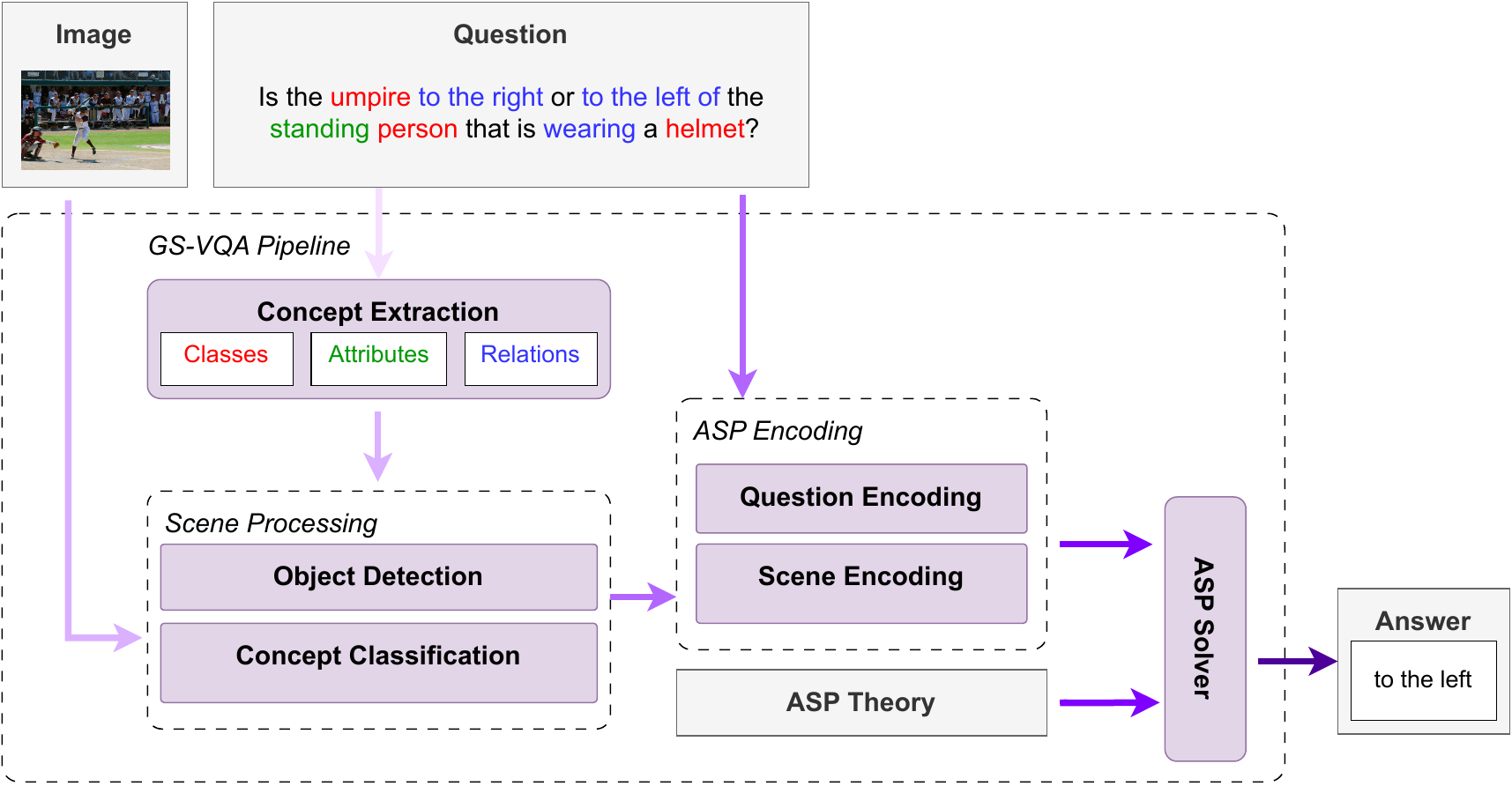}
    \caption{An overview over the full \pipelinetitle pipeline.}
    \label{fig:full_pipeline}
\end{subfigure}%
\begin{subfigure}{.3\textwidth}
  \centering
  \includegraphics[width=1\linewidth]{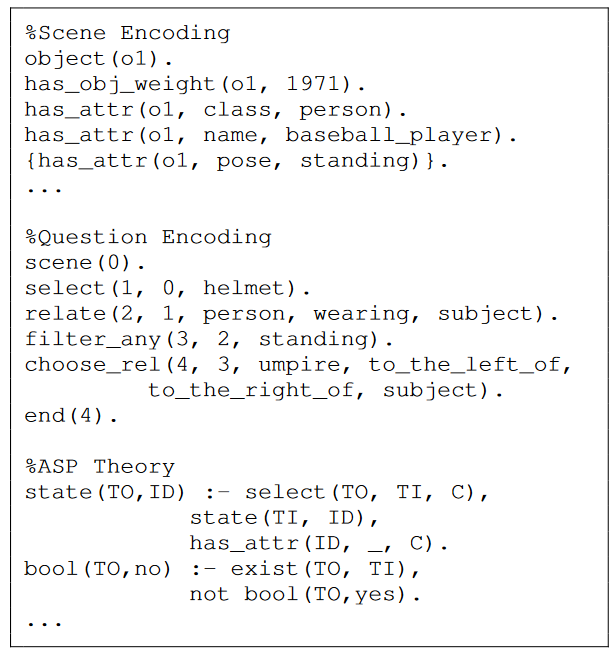}
  \caption{Excerpts from the question encoding, scene encoding, and the ASP theory.}
  \label{fig:encodings}
\end{subfigure}
\caption{\pipelinetitle takes the image and question as input and uses a question-driven approach to generate a partial scene graph. We generate an ASP representation of both partial scene graph and question. These are then solved along an ASP theory to derive the correct answer.}
\end{figure*}

Our system for solving GQA, shown in Fig~\ref{fig:full_pipeline}, resembles other modular neurosymbolic models
consisting of modules for language, vision, and reasoning. We next describe these modules, a performance evaluation is relegated to the appendix.
As mentioned, one of the advantages of using GQA is that we already have a functional representation that we can use for every question.
Neurosymbolic systems have already shown that even classic neural networks such as the LSTM~\cite{hochreiter1997lstm} are able to properly translate natural language into these representations.
Moreover, LLMs can now be used for the task of semantic parsing, which may generalise much better than trained neural networks for specific datasets.
We assume that the representation is given, and thus we only implement a script that translates it to our ASP representation. 
An example of this is shown in Fig.~\ref{fig:encodings} under ``Question Encoding''. The predicates are linked 
by the numbers in the first argument, 
which represent steps to compose questions.

We next discuss the reasoning module, which uses as a fixed logical program.
In a later section, we show how to use LLMs to extend such a logical theory to add functionality.

\smallskip
\noindent
\emph{The ASP theory.$\quad$}
The \pipelinetitle pipeline constructs symbolic encodings of both the input question and the input image, which we call {\em question encoding} and {\em scene encoding}, respectively.
Similarly to a related approach~\cite{eiter2022neuro}, we use ASP as the symbolic formalism for these encodings. This provides us not only with a mature ecosystem of tooling and solvers,
but more importantly, allows us to capture the uncertainty in the class, attribute, and relation predictions of the scene-processing component. 

The ASP theory consists of a set of rules that---in contrast to the question and scene encoding---do not change from question to question and encode the semantics of the reasoning operations that can appear as part of the question encoding.

Revisiting the rules in Figure~\ref{fig:encodings},
there $T_i$ and $T_o$ are variables representing input/output step references, $\mathit{ID}$ represents an object id, $C$ a class, $A$ an attribute category, $V$ an attribute value, and $R$ a relation.
The entire \theory, described as part of the online repository, is solved alongside both scene and question encodings to produce an answer. 

\smallskip
\noindent
\emph{Question-driven scene-graph generation.$\quad$}
The zero-shot nature of the pipeline presents a challenge for constructing a complete scene graph of visually complex scenes: due to the general-purpose nature, inference with the VLMs that \pipelinetitle uses for scene graph generation is far more resource intensive than adhoc trained models.
As such, constructing a full scene graph in which likelihoods for all possible classes, attributes, and relations in the dataset are present for every object detected in the scene is untenable
within a time bound that a human user might find acceptable. 

To resolve this issue, the  pipeline resorts to {\em ``question-driven'' partial scene-graph extraction}, where only information is extracted from the scene that is relevant for answering the question at hand. 
To this end, the {\em concept-extraction component} determines which object classes, attributes, and relations are relevant from the semantic representation of the input question. 
Conceptually, it takes a question in natural language and produces a tuple $(C, A, R)$, where $C$ is a set of classes, $A$ is a set of attribute categories, and $R$ is a set of relations.
We use the functional representation that comes with every question in GQA to accomplish this.
For the example question in Fig.~\ref{fig:full_pipeline}, the tuple is 
$(\{\mathtt{helmet}, \mathtt{umpire}, \mathtt{person}\}, \{\mathtt{pose}\}$, 
$ \{\mathtt{wearing}, \mathtt{to\_the\_left\_of}, \mathtt{to\_the\_right\_of}\})$.

Using the $(C, A, R)$ output tuple of the concept-extraction component, the {\em scene-processing component} has the task of extracting a question-driven partial scene graph. We represent the graph as a list $O=[o_1, \ldots, o_n]$ of objects $o_i = (id, s, B, c, A_o, R_o)$,  each having a unique identifier $id$, a score $s$ between 0 and 1 denoting the confidence in the  object detection from scene-processing, a bounding box $B$, a class $c$, and sets $A_o$ and $B_o$ of attribute and relation likelihoods, respectively. The class $c$ is either in $C$ or a sub-class of maximal specificity of a class $c'$ in $C$. The latter ensures that an object cannot, \egc, be detected as just a ``person'', but must be detected as a maximally specific class like ``baseball player''.  It also means that the scene graph contains only objects of classes that are deemed relevant to answering the question (hence the description as a ``partial'' scene graph). The set $A_o$ contains, 
for each possible value $v$ of each attribute category $a$ in $A$, a likelihood between 0 and 1 that $v$ applies to $a$ for the object. Finally, $R_o$ contains, for each relation $r \in R$ and each other detected object $o_j\neq o_i$,  a likelihood between 0 and 1 that $r(o_i,o_j)$ applies. 
The scene graph $O = [o_1, o_2, \ldots, o_n]$ generated by when processing the scene is easily translated using a script into our ASP representation. Fig.~\ref{fig:encodings} shows an excerpt of such representation under ``Scene Encoding.''

\subsection{VQA for the CLEVR Dataset}

The second dataset we consider is CLEVR~\cite{johnson2017clevr}, which uses synthetic scenes but challenges VQA systems with more complex compositional questions. We use a VQA system for CLEVR~\cite{eiter2022neuro}, that we have extended by an explanation component in recent work~\cite{EiterGHO23}. We revisit its basic functionality in the remainder of this section.

\subsubsection{The CLEVR dataset.} 

CLEVR was designed to test VQA system with compositional questions that involve making several reasoning steps to derive the correct answer.
The dataset contains synthetically generated images with different objects in it.  These objects vary in their shape (cube, cylinder, sphere), 
colour (brown, blue, cyan, gray, green, purple, red, yellow), 
size (big, small), and 
material (metal, rubber).

CLEVR questions require, \egc identifying objects, counting, filtering for attributes, comparing attributes, and spatial reasoning. 
They are formulated in natural language, but, as for GQA, a functional representation is also provided that can be directly parsed into ASP facts.
For illustration,  the question ``How many large things
are either cyan metallic cylinders or yellow blocks?'' 
then becomes 
{
\[
\begin{array}{@{}l}
\mathtt{end(8).}~\mathtt{count(8,7).}~\mathtt{filter\_large(7,6).}~\mathtt{union(6,3,5).} \\
\mathtt{filter\_cylinder (3,2).}~\mathtt{filter\_cyan(2,1).} \\
\mathtt{filter\_metal(1,0).}\\
\mathtt{filter\_cube(5,4).}~\mathtt{filter\_yellow(4,0).}~\mathtt{scene(0).}
\end{array}
\]
}
which encodes an execution tree of operations to derive the answer, where indices refer to output (first argument) and input (remaining arguments) of the respective operations.

\subsubsection{Solving CLEVR.} 

The architecture of the system for CLEVR is similar to the one for solving GQA. In fact, the latter system was inspired by the design of the former. The scenes in CLEVR are however less complex, and we use the 
popular object-detection framework YOLOv5\footnote{\url{https://ultralytics.com/yolov5}.} to
identify all objects in the image, their attributes, and further bounding-box information. 

As for GQA, ASP is used for the reasoning module. More specifically, we use a uniform ASP encoding that describes how to derive the answer for a question and a scene, both given as translations into ASP facts, by step-wise evaluating the execution tree of operations from the question encoding. More information, code, and the full ASP encoding can be found online.\footnote{\url{https://github.com/pudumagico/nsvqasp}.}

\section{Knowledge Distillation Method}\label{sec:distillation}

Our VQA approaches use a hand-coded ASP \theory that correctly computes the answer to any question in the dataset, given a correct representation of both the question and the image. 
The ASP encoding is constructed around a fixed dataset. If the dataset is extended, rules need to  be modified or added to handle new examples.
In general, these new rules must be crafted by a human; here, an LLM based system can come to aid and provide automated support.
To this end, we aim for a reasoning module that manages the \theory by being able to recognise which examples it can handle and which it cannot, and in the latter case, by adding rules in a way such that the example can be solved.
We propose a method of \emph{declarative knowledge distillation}, where the model we distil from is an LLM, and the knowledge that is distilled from it is represented in ASP rules.

\subsection{Preprompt}

We first present a preprompt that instructs the LLM to only return ASP rules that extend an initial \theory $\mathrm{Init}$.
Theory $\mathrm{Init}$ is a partial encoding for the task at hand that we want to extend.
Our preprompt consists of several components:

\smallskip

\begin{compactenum}
    \item {\bf Introduction}: We present the setting of VQA and clarify that we have already parsed both scene and question into correct representations.
    \item {\bf Language Syntax}: We describe the syntax of the language we use to represent questions and scenes, in our case ASP.
    \item {\bf Scene and Question Explanation}: We explain the representation that is used for the scene graph and questions and give  examples.
    \item {\bf Answer Format}: We describe the format of the answers to the questions.
    \item {\bf Initial Theory}: We present the initial \theory $\mathrm{Init}$ that must be updated  as necessary.
    \item {\bf Task Explanation}: Finally, we explain the input the LLM will receive from now on and the expected response (see Listing~\ref{lst:task_preprompt}).
\end{compactenum}
\smallskip

\noindent The input after the preprompt are question/scene/answer tuples $(Q,S,A)$ in the language of ASP. 
The task is the following: when our system recognises that the current \theory cannot handle the instance presented, then the LLM is prompted to add rules 
such that the correct answer can be derived. The expected LLM output is a list of ASP rules.

\smallskip

\begin{lstlisting}[basicstyle=\footnotesize\ttfamily,captionpos=b,caption=Excerpt from the Task Explanation part of our preprompt.,label=lst:task_preprompt,numbers=none]
Your task is to keep the ASP theory 
updated with rules that allows us
to answer questions.
We provide an initial theory that 
can handle some instances.
The prompt input will consist of one or 
more questions in the ASP representation.

Strictly follow these guidelines:
1. Only output the new ASP Rules.
2. Do not add facts as rules.
3. New rules should be as general 
as possible, i.e., have a low number of 
constants and high number of variables.
4. Do not output any natural language.
\end{lstlisting}

\subsection{Rule Distillation Algorithm}

With the task 
explained to the LLM by the preprompt, we can start to present examples from the dataset. 
For each question/scene/answer tuple $p = (Q,S,A)$, we do the following steps (and repeat them for the same example at most $r$ times if not successful):

\smallskip

\begin{compactenum}

\item {\bf Prompting:} We prompt the LLM with $p$ as additional input, and we get as a response $R$.  
\item {\bf Solving:} We concatenate $R$ with the initial \theory to get theory $\mathrm{Res}$. Then, we run an ASP solver on \theory $\mathrm{Res}$ alongside the instance pair $(Q,S)$.

\begin{compactenum}

    \item{\bf Syntax Check:} If we get a syntax error, we pass the error message to the LLM and prompt it to revise $R$ accordingly. We try this at most $m$ times.

    \item {\bf Semantic Check:} We check whether the answer we get from the solver is correct, \iec coincides with $A$. 
    If not, we pass the actual answer 
    and the expected answer to the LLM and task it to update $R$; we try this at most $m$ times. 

\end{compactenum}

    \item {\bf Regression Testing:} To avoid that adding rules to the theory renders past examples incorrect, we test $\mathrm{Res}$ on all previously seen examples. Only if all tests pass we keep $\mathrm{Res}$, otherwise we disregard the extension $R$.

\end{compactenum}

\smallskip
The algorithm has two parameters, $r$ is the number of retries per example, and $m$ is the number of retries for mending incorrect rules (defaults are $r=m=1$).
Mending rules is potentially expensive as it requires more calls to the LLM; it can be turned off if preferred.

\subsubsection{Example Sampling Strategies.} 

VQA datasets contain millions of instances, and going blindly through them can make the distillation process ineffective.
Choosing a small but representative sample can yield better results faster.
We propose two strategies to group instances:
\smallskip
\begin{compactitem}
    \item {\bf\em Predicate Count}: We group all the instances by the number of predicate occurrences that appear in the ASP question representation. For this, we create a dictionary whose keys are the numbers of predicates and the contents are questions with such length. 
   \item{\bf\em Predicate Relevance}: Here, we group examples based on the predicates that appear in the question representation. We first create a dictionary whose keys are all the predicates that appear in any question representation. Then we populate the dictionary with questions where the key predicate appears in the question representation. 
   
\end{compactitem}
We then sample examples from the group created for the chosen strategy with a parameter $k$.
For the former, the total number of examples is $k$ multiplied by the number of keys in the dictionary, while for the latter, the total number of examples is exactly $k$.

\subsubsection{Batch Optimization.} 

We present prompt instances one by one, which results in one LLM call for each example. This is not very efficient and can misguide the LLM into implementing rules that only solve that particular example, yet our aim is to have general rules that can handle a considerable portion of the dataset.
Considering that LLMs have increasingly larger context sizes, we also investigate the option of presenting prompt instances in batches, where each batch contains up to $b$ singular instances.
We observe that the scene representation is usually much larger than the question representation and only contains a small number of predicates, and the variance comes more from the constants in them.
Considering this, whenever we use our batch strategy, we present the prompt instance $p = \{Q_1,\ldots,Q_b\}$, where we include only the question representation in the $Q_i$ and drop the scene representations.
Now the LLM must create rules general enough to pass the semantic check for all the examples in the batch.
However, there is an expected trade-off between batch size and accuracy: With large $b$, the semantic check and regression testing might be too strict for the LLM to produce rules that correctly cover all examples in one shot.

\section{Knowledge Distillation Experiments}\label{sec:experiments}

\begin{table*}[!t]
\centering\small
\begin{tabular}{lrr@{~\,}lr@{~\,}lr@{~\,}l}
\toprule
$P$     & $\mathrm{Init} \setminus P$ & \multicolumn{2}{c}{GPT-4} & \multicolumn{2}{c}{GPT-3.5} & \multicolumn{2}{c}{Mistral}\\ 
\midrule
query   & $48.84$ & $97.67\pm18.05$ & $(89.16, 98.92)$ &$70.02\pm19.36$ & $(48.84, 85.53)$ & \multicolumn{2}{c}{---}\\ 
exist   & $86.36$ & $99.75\pm00.50$ & $(98.86, 99.98)$& $87.65\pm02.41$ & $(86.36, 91.95)$ & $89.68\pm04.20$ & $(86.36, 95.66)$\\ 
or      & $92.18$ & $100.0\pm00.00$ & $ (100.0, 100.0)$& $93.03\pm01.90$ & $(92.18, 96.44)$ & $93.20\pm01.66$ & $(92.18, 96.02)$\\ 
filter  & $81.60$ & $98.21\pm00.40$ & $(97.49, 98.40)$ & $83.15\pm03.47$ & $(81.60, 89.37)$  & $81.70\pm00.24$ & $(81.60, 82.14)$\\ 
choose\_attr & $92.12$ & $95.98\pm05.37$ & $(88.73, 99.83)$& $93.73\pm01.36$ & $(92.31, 95.83)$ & $92.12\pm00.01$ & $(92.12, 92.15)$\\ 
verify\_rel & $93.72$ & $98.60\pm01.11$ & $(96.73, 99.43)$& \multicolumn{2}{c}{---} &  \multicolumn{2}{c}{---}\\ 
select & $9.53$ & $99.94\pm00.07$ & $(99.87, 100.0)$ & $27.42\pm40.01$ & $(9.53,99.01)$ & \multicolumn{2}{c}{---}\\ 
negate & $98.59$ & $98.54\pm00.20$ & $(98.59, 98.74)$ & \multicolumn{2}{c}{---} & \multicolumn{2}{c}{---}\\ 
relate & $56.89$ & $69.38\pm12.50$ & $(56.89, 85.25)$ & \multicolumn{2}{c}{---} & \multicolumn{2}{c}{---}\\ 
two\_different & $98.94$ & $100.0\pm00.00$ & $(100.0, 100.0)$ & $99.39\pm00.55$ & $(98.94, 100.0)$ & \multicolumn{2}{c}{---}\\ 
two\_same & $98.83$ & $99.99\pm00.00$ & $(99.99, 100.0)$ &$99.05\pm00.53$ & $(98.83,100.0)$ & \multicolumn{2}{c}{---}\\ 
\bottomrule
\end{tabular}

\smallskip
{\centering (a) Results for GQA.}
\medskip


\begin{tabular}{lrr@{~\,}lr@{~\,}lr@{~\,}l}
\toprule
$P$     & $\mathrm{Init} \setminus P$ & \multicolumn{2}{c}{GPT-4} & \multicolumn{2}{c}{GPT-3.5} & \multicolumn{2}{c}{Mistral}\\ 
\midrule
exist           & $79.63$ & $99.48\pm00.70$ & $(98.72, 100.0)$ & $87.82\pm11.11$ & $(98.72, 100.0)$ & $80.21\pm06.36$ & $(72.16, 90.02)$\\ 
unique          & $29.19$ & $97.67\pm18.05$ & $(89.16, 98.92)$ & \multicolumn{2}{c}{---}            & \multicolumn{2}{c}{---}\\ 
count           & $98.01$ & $99.60\pm00.88$ & $(98.01, 100.0)$ & $98.01\pm01.12$ & $(98.01, 98.40)$ & \multicolumn{2}{c}{---}\\ 
equal\_integer  & $96.61$ & $99.92\pm00.17$ & $(99.61, 100.0)$ & $97.80\pm01.26$ & $(96.61, 99.60)$ & \multicolumn{2}{c}{---}\\ 
and             & $93.67$ & $100.0\pm00.00$ & $(100.0, 100.0)$ & $97.46\pm03.46$ & $(93.67, 100.0)$ & \multicolumn{2}{c}{---}\\ 
relate\_left    & $84.73$ & $100.0\pm00.00$ & $(100.0, 100.0)$ & $96.18\pm07.63$ & $(84.73, 100.0)$ & $94.14\pm08.02$ & $(84.73, 100.0)$\\ 
filter\_large   & $68.54$ & $100.0\pm00.00$ & $(100.0, 100.0)$ & $87.41\pm17.23$ & $(68.54, 100.0)$ & $81.12\pm17.23$ & $(68.54, 100.0)$\\ 
query\_shape    & $72.23$ & $100.0\pm00.00$ & $(100.0, 100.0)$ & $100.0\pm00.00$ & $(100.0, 100.0)$ & $94.44\pm12.43$ & $(72.23, 100.0)$\\ 
same\_color     & $94.79$ & $99.36\pm00.87$ & $(98.41, 100.0)$ & $100.0\pm00.00$ & $(100.0, 100.0)$ & $97.07\pm02.70$ & $(94.79, 100.0)$\\ 
\bottomrule
\end{tabular}

\smallskip
{\centering (b) Results for CLEVR.}

\caption{Results for the knowledge distillation method when attempting to restore $\mathrm{Init}$ after all rules that mention a predicate $P$ are removed.}
\label{tab:predicate_removal_results}
\end{table*}


\begin{table*}[t!]
\centering

\begin{tabular}{cc}

\begin{minipage}{.45\textwidth}
\centering
{\small
\begin{tabular}{lrcc}
\toprule
 $s (\%)$  & Init & GPT-4 & GPT-3.5\\ 
\midrule
10 & 
$26.57$ & \makecell{$94.67\pm02.21$\\$(89.71, 95.67)$} & \multicolumn{1}{c}{---}    \\ 
\midrule    
20 & 
$63.54$ & \makecell{$75.56\pm11.86$\\$(63.55, 90.22)$} & \makecell{$66.14\pm07.78$\\$(63.55, 89.48)$}\\ 
\midrule
50 & 
$7.17$ & \makecell{$47.48\pm15.26$\\$(30.25, 71.64)$} & \makecell{$24.43\pm12.28$\\$(07.18, 46.88)$}\\ 
\bottomrule
\end{tabular}

\smallskip
(a) Results for GQA.

}
\end{minipage}

& 

\begin{minipage}{.45\textwidth}
\centering
{\small
\begin{tabular}{lrcc}
\toprule
 $s (\%)$  & Init & GPT-4 & GPT-3.5\\ 
\midrule
10 & 
$46.61$ & \makecell{$70.76\pm04.17$\\$(66.59, 75.62)$} & \makecell{$50.30\pm03.68$\\$(46.62, 53.98)$}  \\ 
\midrule
20 & 
$9.66$ & \makecell{$44.66\pm30.58$\\$(14.16, 97.30)$} & \makecell{$32.03\pm11.86$\\$(13.44, 45.70)$}\\ 
\midrule
50 & 
$0.0$ & \makecell{$23.90\pm03.30$\\$(18.06, 27.84)$} & \makecell{$10.19\pm19.59$\\$(00.00, 49.36)$}\\ 
\bottomrule
\end{tabular}

\smallskip
(b) Results for CLEVR.

}
\end{minipage}

\end{tabular}

\caption{Knowledge distillation results when attempting to restore a complete ASP theory after a percentage $s$ of rules is randomly removed.}
\label{tab:random_removal_results}
\end{table*}


\begin{table*}[t!]

\begin{tabular}{cc}

\begin{minipage}{.45\textwidth}
    
\small
\centering
\begin{tabular}{lccc}
\toprule
  $b$   & 
        $\mathrm{Light}$ & $\mathrm{Medium}$ & $\mathrm{Heavy}$ \\ \midrule
Init       & 
        $0.0$ & $0.0$ &  $6.24$   \\ \midrule
1       & 
        \makecell{$56,26\pm10.23$\\$(34.54, 61.28)$} & \makecell{$81.45\pm05.07$\\$(76.86, 87.91)$}  &  \makecell{$83.85\pm02.49$\\$(81.38, 87.77)$}   \\ \midrule
2       & 
        \makecell{$32.71\pm04.31$\\$(25.72, 43.15)$} & \makecell{$79.83\pm03.42$\\$(75.11, 83.03)$}  &  \makecell{$74.32\pm02.91$\\$(75.86, 80.54)$}   \\ \midrule
5       & 
        \makecell{$16.62\pm05.28$\\$(10.51, 17.59)$} & \makecell{$69.68 \pm31.12$\\$(24.18,82.19)$}   &  \makecell{$84.25\pm04.59$\\$(78.93,89.48)$}  \\ \midrule
10      & 
               \multicolumn{1}{c}{---}               &  \makecell{$15.38\pm12.30$\\$(11.62, 31.75)$}  &  \makecell{$84.75\pm04.20$\\$(80.64,90.85)$}   \\ \bottomrule

\end{tabular}

\smallskip
(a) Results for GQA.

\end{minipage}

&

\begin{minipage}{.45\textwidth}

\small
\centering
\begin{tabular}{lccc}
\toprule
  $b$   & 
        $\mathrm{Light}$ & $\mathrm{Medium}$ & $\mathrm{Heavy}$ \\ \midrule
Init       & 
        $0.0$ & $5.56$  &  $20.80$   \\ \midrule
1       & 
        \makecell{$84.68\pm26.42$\\$(38.23, 100.0)$} & \makecell{$86.97\pm04.35$\\$(83.89, 90.05)$}  &  \makecell{$95.40\pm03.83$\\$(91.25, 98.81)$}   \\ \midrule
2       & 
        \makecell{$75.4\pm33.78$\\$(27.84, 99.88)$} & \makecell{$18.68\pm04.50$\\$(15.67, 26.09)$}  &  \makecell{$88.51\pm04.46$\\$(83.37, 91.25)$}   \\ \midrule
5       & 
        \makecell{$17.06\pm29.55$\\$(00.00, 51.19)$} & \makecell{$17.79\pm03.00$\\$(15.67, 19.92)$}   &  \makecell{$94.39\pm03.71$\\$(91.33,98.52)$}  \\ \midrule
10      & 
               \multicolumn{1}{c}{---}               &  \multicolumn{1}{c}{---}                     &  \makecell{$89.88\pm09.04$\\$(77.68,98.81)$}   \\ \bottomrule

\end{tabular}

\smallskip
(b) Results for CLEVR.

\end{minipage}

\end{tabular}

\caption{Results for the knowledge distillation method when using batch sizes $b$ and the different initial theories $\mathrm{Light}$, $\mathrm{Medium}$ and $\mathrm{Heavy}$.}

\label{tab:batch_results}
\end{table*}

\begin{table*}[!t]
\centering\small
\begin{tabular}{lrr@{~\,}lr@{~\,}lr@{~\,}l}
\toprule
$P$     & $\mathrm{Init} \setminus P$ & \multicolumn{2}{c}{GPT-4} & \multicolumn{2}{c}{GPT-3.5} & \multicolumn{2}{c}{Mistral}\\ 
\midrule
query   & $48.84$ & $99.02\pm00.04$ & $(99.02, 99.03)$ &$55.90\pm15.80$ & $(48.84, 84.17)$ & \multicolumn{2}{c}{---}\\ 
exist   & $86.36$ & $99.09\pm01.76$ & $(95.94, 99.97)$& $87.83\pm03.29$ & $(86.36, 93.73)$ & $86.60\pm0.05$ & $(86.36, 87.57)$\\ 
or      & $92.18$ & $100.0\pm00.00$ & $ (100.0, 100.0)$& $93.03\pm01.90$ & $(92.18, 96.44)$ & $94.41\pm3.35$ & $(92.18, 99.73)$\\ 
filter  & $81.60$ & $98.56\pm00.58$ & $(98.47, 98.63)$ & $82.38\pm0.078$ & $(81.92, 82.50)$  & $84.99\pm7.59$ & $(81.60, 98.59)$\\ 
choose\_attr & $92.12$ & $98.65\pm02.65$ & $(93.91, 99.88)$& $95.16\pm04.26$ & $(92.03, 99.84)$ & $92.08\pm0.06$ & $(91.98, 92.12)$\\ 
verify\_rel & $93.72$ & $95.74\pm03.49$ & $(93.72, 99.08)$& $94.30\pm01.17$ & $(93.72, 96.06)$ &  \multicolumn{2}{c}{---}\\ 
select & $9.53$ & $81.69\pm36.81$ & $(9.53, 100.0)$ & $28.94\pm39.82$ & $(9.53,100.0)$ & \multicolumn{2}{c}{---}\\ 
negate & $98.59$ & $98.72\pm00.02$ & $(98.59, 99.12)$ & \multicolumn{2}{c}{---} & \multicolumn{2}{c}{---}\\ 
relate & $56.89$ & $58.02\pm02.19$ & $(57.54, 61.91)$ & $57.29\pm00.09 $ & $(56.89, 58.91)$ & \multicolumn{2}{c}{---}\\ 
two\_different & $98.94$ & $100.0\pm00.00$ & $(100.0, 100.0)$ & \multicolumn{2}{c}{---} & \multicolumn{2}{c}{---}\\ 
two\_same & $98.83$ & $99.60\pm00.02$ & $(99.50, 100.0)$ &$99.09\pm00.05$ & $(98.83,100.0)$ & \multicolumn{2}{c}{---}\\ 
\bottomrule
\end{tabular}
\caption{Ablation study for GQA: Results when attempting to restore $\mathrm{Init}$ after removing rules for $P$ without mending step.}
\label{tab:ablation_GQA_predicate_removal_results}

\end{table*}

\begin{table}[!t]
\centering
{\small
\begin{tabular}{lrcc}
\toprule
 $s (\%)$  & Init & GPT-4 & GPT-3.5\\ 
\midrule
10 & 
$26.57$ & \makecell{$83.06\pm23.26$\\$(36.61, 95.80)$} & \multicolumn{1}{c}{---}    \\ 
\midrule
20 & 
$63.54$ & \makecell{$55.46\pm14.84$\\$(26.99, 69.84)$} & \makecell{$38.67\pm21.00$\\$(08.48, 64.96)$}\\ 
\midrule
50 & 
$7.17$ & \makecell{$36.38\pm10.62$\\$(18.15, 48.55)$} & \makecell{$04.21\pm03.42$\\$(00.00, 7.94)$}\\ 
\bottomrule
\end{tabular}
}
\caption{Ablation study for GQA: Attempting to restore theory $T$ after $s$ percent of rules were randomly removed without mending.}
\label{tab:ablation_random_removal_results}
\end{table}

\begin{table}[t!]
\small
\centering
\begin{tabular}{lccc}
\toprule
  $b$   & 
        $\mathrm{Light}$ & $\mathrm{Medium}$ & $\mathrm{Heavy}$ \\ \midrule
Init       & 
        $0.0$ & $0.0$  &  $6.24$   \\ \midrule

1       & 
        \makecell{$51,43\pm08.56$\\$(42.72, 59.85)$} & \makecell{$80.41\pm05.34$\\$(74.54, 85.01)$}  &  \makecell{$76.70\pm00.76$\\$(75.82, 77.19)$}   \\ \midrule
2       & 
        \makecell{$27.06\pm09.60$\\$(21.09, 38.14)$} & \makecell{$77.70\pm05.42$\\$(75.25, 83.92)$}  &  \makecell{$77.60\pm03.85$\\$(73.41, 81.00)$}   \\ \midrule
5       & 
        \makecell{$15.23\pm02.28$\\$(12.60, 16.49)$} & \makecell{$60.16\pm34.17$\\$(21.08,84.45)$} &  \makecell{$86.28\pm10.84$\\$(27.93,94.71)$}  \\ \midrule
10      & 
               \multicolumn{1}{c}{---}             & \makecell{$19.55\pm07.41$\\$(13.03, 27.62)$}  &  \makecell{$73.93\pm06.50$\\$(66.42,77.69)$}   \\ \bottomrule

\end{tabular}
\caption{Ablation study for GQA: Results when using batch sizes $b$ and different initial theories without mending step.}
\label{tab:ablation_batch_results}
\end{table}


\noindent
We conduct a series of experiments to evaluate our knowledge distillation method on GQA and CLEVR to answer the following research questions:%
\footnote{The code for reproducing our experiments is available as an online repository: \url{https://github.com/pudumagico/KR2024}.} 

\smallskip
\noindent
{\em\bf (R1)}
Given an ASP-based VQA system and a VQA task, can our approach extend the ASP reasoning component to deal with questions that require reasoning operations/steps that are not yet implemented?

\smallskip
\noindent
{\em\bf (R2)}
What LLMs are suitable for our method? 

\smallskip
\noindent
{\em\bf (R3)}
Can our method cope with the challenge of removing, either randomly or in a more controlled way,  increasingly large parts from an initial complete theory?

\smallskip
\noindent
{\em\bf (R4)}
What are the effects of the more resource-friendly batch processing variant and mending switched off for our method?

\smallskip

The evaluation platform is a workstation with an Intel Core i7-12700K CPU, 32GB of RAM, and an NVIDIA GeForce RTX 3080 Ti GPU with 12GB of video memory.
All experiments were run $5$ times. 
We include average accuracy, standard deviation, as well as the minimum and maximum value obtained.
For reproducibility, we logged all our parameters, random seeds, and input prompts.

\myparagraph{LLM selection.} 
Before going into the details of our knowledge distillation experiments, we describe how we selected LLMs for further consideration.

We ran the ASP programs of our VQA systems on examples from GQA and CLEVR. 
Then we selected ca.\ $45k$ examples where the answer was correctly calculated for GQA, and then divided this set into a training and a test suite of ca.\ $35k$ and $10k$ instances, respectively. For CLEVR, we selected $50k$ examples and split them into $35k$ for training and $15k$ for testing.\footnote{The expression ``training set'' refers here to the examples used when running the knowledge distillation method.} 

For the selected examples, we ran preliminary experiments on a large array of LLMs, both local and online API-based ones.
Local models, such as GPT4ALL~\cite{DBLP:journals/corr/abs-2311-04931} ``wizardlm-13b'', showed very poor performance when prompted to produce ASP rules.
Some API models, like ``mistral-medium'',  where too slow in coming up with a response for our purposes and were thus excluded.

We selected the three top performers, which were GPT-4~\cite{openai2023gpt4}, GPT-3.5, and Mistral~\cite{jiang2023mistral}; more specifically, the models used are ``gpt-4-1106-preview'', ``gpt-3.5-turbo-1106'' and ``mistral-small'', respectively.

\myparagraph{Experiment~1.} Our first experiment is designed to address {\em\bf (R1)} and {\em\bf (R2)}. 
We start with the complete ASP encoding for each VQA approach presented in Section~\ref{sec:gsvqa}
and remove all rules that mention a particular predicate $P$ that occurs in some question representation. We then prompt the LLM with examples that contain $P$ in their question representation to repair the \theory. This simulates a scenario where the initial ASP theory is not yet capable to address all requirements of the VQA task and needs to be extended.

We use the predicate ordering strategy with a sample size of $k=10$. As we only sample questions where $P$ is present, the number of examples used for each run is~$k$.

The results for GQA and CLEVR are shown in Table~\ref{tab:predicate_removal_results}.
Column $\mathrm{Init} \setminus P$ shows the accuracy of the initial ASP encoding with a predicate $P$ removed. 
The drop in accuracy varies depending on the number of questions affected and the role of the predicate that was removed (\egc ``select'' results in a deeper drop).
The other columns show the performance of the considered LLMs; 
here and in other tables, ``---'' means no improvement.
By a large margin, GPT-4 is the most suitable LLM for this task for both datasets.
With GPT-4, we could obtain rules that improve over $\mathrm{Init}$ for every predicate $P$. For the other models, there the quality of suggested repairs differs largely. GPT-3.5 is also capable of repairing $\mathrm{Init}$ but to a lesser extent than GPT-4, as it does not produce any new rules for some predicates. When it does, the accuracy is lower than the one of GPT-4.
The Mistral model performs similarly to GPT-3.5 when it finds the correct rules. This, however, happened less often and the gain in accuracy then is minimal.

\myparagraph{Experiment~2.} This experiment is designed to address {\em\bf (R3)}. 
It is a challenge experiment where we took the complete ASP encodings and then removed a random sample of $s$ percent of the rules from it.
For GQA, the ASP theory consists of 60 rules; for CLEVR, the program comprises 72 rules. 
After rendering this theory incomplete, we prompted the LLM to restore it. 
We tested this setting for $s=\{10,20,50\}$ on the two best models from the first experiment, \iec GPT-4 and GPT-3.5. We used the length ordering strategy with $k=2$, which yields $20$ examples per run, and $r=1$ retries.

The results are given in Table~\ref{tab:random_removal_results}. For GQA, GPT-4 is quite capable of producing ASP rules that improve the accuracy of the initial theory.
However, GPT-3.5 starts to falter for GQA, as the gain in accuracy is dramatically reduced for $s=20,50$, and no gain is reported when $s=10$.
In the case of GQA, the initial accuracy of the theory with $s=10$ is higher than for $s=20$. Even though we remove more rules in the latter, some rules affect larger portions of the questions.

\myparagraph{Experiment~3.} This experiment tests our batch optimization approach and aims at answering {\em\bf (R4)}. We consider batch sizes $b={2,5,10}$ and
the length ordering strategy with $k=11$, which yields $100$ examples per run. We use $r=3$ retries for examples and $m=2$ retries for mending.

Like the previous experiment, this one is also based on the complete ASP theory, but we now selected parts of it by hand to generate three different initial theories.
By their size, we call them $\mathrm{Light}$, $\mathrm{Medium}$ and $\mathrm{Heavy}$. Theory $\mathrm{Light}$ consists of only five rules that tell our system how to start and end the processing of a question, but nothing else.
We added more rules to $\mathrm{Light}$ to generate $\mathrm{Medium}$ and even more rules to $\mathrm{Heavy}$. For this experiment, we focused only on the most capable model, namely GPT-4.

Table~\ref{tab:batch_results} shows that GPT-4 still retains its ability to produce meaningful ASP rules even in this very challenging scenario.
We can observe that small theories do not work well with bigger batches, as they do not provide enough background for the LLM to produce suitable rules.
When the size of the initial theory increases, the LLM can handle bigger batches and produces rules of higher quality.

\myparagraph{Ablation Experiment.}
We finally study the effect of the mending step in our method when the LLM suggests rules that are either syntactically or semantically not correct. As mentioned in the previous section, these checks can be turned off, which results in fewer LLM calls.

We only present the results  for the previous three experiments on GQA in Tables~\ref{tab:ablation_GQA_predicate_removal_results}--\ref{tab:ablation_batch_results}; for CLEVR, they look similar.
As one would expect, the results are worse most of the time, although not always, when mending is disabled. The rate of improvement with mending ranges from a couple of percentage points to up to ca.\ $20\%$ (row for ``select'').

\myparagraph{Discussion.} 
We turn back to our research questions from the beginning of this section.

\smallskip
\noindent
{\bf (R1)} 
Experiment~1 shows that LLMs are capable of completing an ASP program that has all rules for a single operation removed. This is the case when a dataset is extended with new examples that require reasoning operations that are not yet encoded. 

\smallskip
\noindent
{\bf (R2)}
Regarding the suitability of different LLMs for declarative knowledge distillation, we  conclude that only grand-scale LLMs, with GPT-4 currently the market leader, are able to tackle this problem effectively. Arguably, the LLMs that we used have more knowledge about mainstream programming languages such as Python than logical programming languages. It will be interesting so see whether small, self-hosted language models 
will eventually catch up in the future. 

\smallskip
\noindent
{\bf (R3)} 
When challenged with restoring increasingly large parts of an ASP theory, the current approach reaches its limits. Only the most powerful model we used is still able to produce sound ASP rules.

\smallskip
\noindent
{\bf (R4)}
Our experiments on batch processing and the ablation study helped to illuminate the trade off between performance and reducing the number LLM calls.
In conclusion, better performance can be bought by using more prompts, which can be expensive if a subscription-based LLM is used. 

\section{Conclusion}\label{sec:concl}

We have presented an approach for declarative knowledge distillation 
using LLMs to find rules that extend the reasoning component of a VQA system to extend its capabilities. This process uses examples from a dataset as guidance and relies entirely on prompting without the need to train or fine-tune the used language models. We have demonstrated this approach on the prominent CLEVR~\cite{johnson2017clevr} and GQA~\cite{hudson2019gqa} datasets. The benefit of using logic-based methods in combination with foundation models is that we obtain systems that behave in an interpretable and verifiable way to ensure correct reasoning. Logical rules are intuitive, 
and they can helpful in VQA architectures to create advanced reasoning capabilities, including explainability. Our knowledge distillation method shows promise for automating the process of ASP modelling for VQA, a complex scenario that requires understanding of intricate representations.

For future work, we want to study further VQA datasets, but we also want to explore other tasks than VQA that potentially benefit from distilling rules from LLMs.
As using API-based LLMs can be expensive, 
we want to look into balancing performance of the distillation approach and the number of LLM calls.
Enhancing the LLM approach with concepts from ILP and a possible combination would also be interesting to explore.

\appendix
\section*{Appendix}

We provide an evaluation of our logic-based VQA approach, which we call \pipelinetitle.
We evaluated the \pipelinetitle pipeline on the balanced test-dev set of GQA, which contains $12\,578$ questions.
The evaluation uses the larger ViT-L/14 variant of OWL-ViT~\cite{minderer2022owlvit} for object detection and the smaller ViT-B/32 variant of CLIP \cite{radford2021clip} for concept classification.

\begin{table}[t!]
    \centering
    \begin{tabular}{lllr}
        \toprule
        & {Model} & {Category} & {Accuracy} \\
        \midrule
        \multirow{6}{*}{} 
        & BLIP-2 & end-to-end & $44.7\%$ \\
        & CodeVQA & question-symbolic & ${49.0\%}$ \\ 
        & FewVLM  & end-to-end & $29.3\%$ \\
        & \pipelinetitle $\text{(ours)}$ & neurosymbolic & $39.5\%$ \\ 
        & PnP-VQA  & semi-symbolic & $42.3\%$ \\
        & ViperGPT & question-symbolic & $48.1\%$ \\
        \bottomrule
    \end{tabular}
    \caption{Comparison of \pipelinetitle's accuracy on the test-dev set of GQA with other state-of-the-art zero-shot approaches for VQA.}
    \label{tab:pipeline_results}
\end{table}

In Table~\ref{tab:pipeline_results}, we present a comparison between our system and other models on GQA. 
We consider only zero-shot models and classify them into four sub-categories:
\begin{compactitem}
\itemsep=0pt
    \item {\bf End-to-End}: 
End-to-end systems are those that rely solely on neural networks for  computing the answer.
\item {\bf Neurosymbolic}: Neurosymbolic systems like ours are those that combine both neural networks for parsing data and symbolic execution to calculate the answers.
\item {\bf Question-Symbolic}: 
Such methods extract a symbolic representation from only the input question, usually in the form of some programmatic specification of the reasoning steps needed to arrive at the answer of the question.
\item {\bf Semi-Symbolic}: PnP-VQA~\cite{tiong2023pnpvqa} extracts a symbolic representation of the image but does not perform its reasoning purely symbolically, hence 
we classify this method as semi-symbolic.
\end{compactitem}

\pipelinetitle answers $39.5\%$ of all questions of GQA's test-dev set correctly, with the current best zero-shot VQA model, CodeVQA \cite{subramanian2023codevqa}, obtaining an accuracy of $49.0\%$, followed closely by ViperGPT \cite{vipergpt}. 
However, CodeVQA and ViperGPT translate input questions into Python code that may contain queries to another VQA model;
thus the performance of the 
latter, which is PnP-VQA \cite{tiong2023pnpvqa} 
resp.\ BLIP-2 \cite{li2023blip2}, should be considered as their baseline.

While our model is slightly behind the 
baseline in terms of accuracy by a couple of percentage points, it uses ASP for deducing answers which comes with a range of advantages, where explainability and non-determinism to deal with ambiguous inputs are the most important ones \cite{eiter2022neuro}. 
More specifically, ASP allows for transparent execution, verifiability, and transferability, as well as the possibility of reasoning under different modalities, \egc abduction on instances for computing contrastive explanations~\cite{EiterGHO23}.
Furthermore, the components used for our system were run locally, in contrast to others such as ViperGPT which hinge on extensive external resources.
Finally, since our architecture is modular, components can be easily replaced by better ones, and it can be expected that its performance improves when the underlying foundation models get better.

\paragraph{Acknowledgment.} 
This work was supported by the Bosch Center for AI.

\bibliographystyle{kr}
\bibliography{refs}

\end{document}